\documentclass[12pt]{article}
\usepackage{fullpage}
\usepackage{hyperref}
\usepackage{graphicx}
\usepackage[export]{adjustbox}
\usepackage{array}
\usepackage{listings}
\usepackage[utf8]{inputenc}
\newcommand{\realization}[1]{\textsf{#1}}

\newcommand{\systeme}[1]{\textsc{#1}}
\newcommand{\jsr}{\systeme{jsRealB}}
\newcommand{\js}{\systeme{JavaScript}}
\newcommand{\py}{\systeme{Python}}
\newcommand{\pyr}{\systeme{pyrealb}}
\newcommand{\snlg}{\systeme{SimpleNLG}}
\newcommand{\kpml}{\systeme{KPML}}
\newcommand{\surge}{\systeme{Surge}}
\newcommand{\realpro}{\systeme{RealPro}}

\newcommand{\gscale}{0.5}

\title{The \jsr{} Text Realizer:\\ Organization and Use Cases\\ \emph{Revised version}}

\author{
Guy Lapalme\\
RALI-DIRO\\ 
\url{lapalme@iro.umontreal.ca}\\
}

\begin{document}
\maketitle
\begin{abstract}
This paper describes the design principles behind  \jsr{} (Version 4.0), a surface realizer written \js{} for English or French sentences from a specification inspired by the \emph{constituent syntax} formalism but for which a dependency-based input notation is also available. \jsr{} can be used either within a web page or as a \lstinline!node.js! module. We show that the seemingly simple process of text realization involves many interesting implementation challenges in order to take into account the specifics of each language. \jsr{} has a large coverage of English and French and has been used to develop realistic data-to-text applications and to reproduce existing literary texts and sentences from Universal Dependency annotations. Its source code and that of its applications are available on GitHub. The port of this approach to Python (\pyr{}) is also presented.
\end{abstract}

\section{Introduction} 
\label{sec:introduction}

A text realizer sits at the very end of a text generation pipeline. Important decisions about \emph{What to say} must have already been made, sentence structure and most content word choices must also have been decided. Final realization is an often \emph{neglected} part in NLG systems because it is dubbed to be \emph{pedestrian}, often associated with \emph{glorified} format statements, even though its output is the only thing that the end user sees and that is used to evaluate the whole system. How reasonable is an output if word agreements are not properly done or if it consists of a mere list of tokens?  This might be sufficient for machine evaluation, but it cannot practically be used in a production setting.  A well formatted and grammatically correct output is important for the \emph{social acceptability} of a system. This is why we decided to study this process more closely.

A text realizer has much interesting work to do: it must take care of many language peculiarities such as number and person agreements, conjugation, word order and elision. On top of these tasks, \jsr{} also allows creating many variations (e.g.,~negative, passive or interrogative) from a single affirmative sentence structure. We first briefly present \jsr{} and compare it with existing text realizers.


\section{\jsr{}} 
\label{sec:_jsr}

\jsr{} (\js{} Realizer Bilingual)~\cite{JSreal14,molins-lapalme-2015-jsrealb} is a bilingual French and English text realizer that generates well-formed expressions and sentences and that can format them in HTML to be displayed in a browser. As its name indicates, \jsr{} is written in \js{}, a programming language that, when used in a web page, runs in the client browser. A web programmer who wishes to use \jsr{} to produce flexible French or English textual output only needs to add one line in the header of the page, similarly to what is done for other browser frameworks such as \systeme{jQuery}. \jsr{} is aimed at web developers. It carries out crucial tasks from taking care of morphology, subject-verb agreement and conjugation to creating entire HTML documents. \jsr{} has been used for creating data-to-text systems. It has also been used as an intermediary for realizing sentences produced by a Prolog program taking input from AMR structures~\cite{AMR-2019} or from Universal Dependencies in the context of the Surface Realization Shared Task (SR'19) at EMNLP~\cite{Lapalme19SRST} and for the WebNLG Challenge~\cite{castro-ferreira20:_2020_bilin_bidir_webnl}.

\jsr{} was strongly influenced by \snlg{}~\cite{gatt-reiter-2009-simplenlg} in which realization is achieved by programming language instructions that create data structures corresponding to the constituents of the sentence to be produced. Once the data structure (a tree) is built in memory, it is traversed to produce a string. Like \snlg{}, \jsr{} has the following components :
\begin{itemize}
    \item a lexicon defining the word category, gender, number, irregularities and other features;
    \item morphological rules to determine the appropriate word forms, such as plurals and conjugations;
    \item syntactic rules to properly order words in a sentence, perform agreement between constituents and carry out other interactions.
\end{itemize}

\jsr{} integrates other useful tools, such as the spelling out of numbers and the wording of temporal expressions. Since it produces web content, it can add HTML tags for links, for formatting, to create headers and lists. It takes them into account so that they do not interfere with the proper processing of words within a sentence. 

\jsr{} has been under development in our lab since 2013\footnote{The source code is available on the RALI-\href{https://github.com/rali-udem/jsRealB}{GitHub} with a tutorial and many demonstration applications.}. Since then it has been reengineered while keeping the same external interface. This paper is the result of this recent work, which revealed many intricacies that must be dealt with and that we thought it would be interesting to document more formally. In the current version 4.0, a dependency notation has been added. \jsr{} has also been ported to Python keeping the same notation for specifying the content of the sentences, see Section~\ref{sub:python_implementation} for more details.

\subsection{Input to the realizer} 
\label{sec:input_to_the_realizer}

In \jsr{}, \js{} expressions create data structures corresponding to the constituents of the sentence to produce. When the need arises to produce a string realization (i.e. its \lstinline!toString()! function is called), the data structure (a tree) is traversed to produce the tokens of the sentence, taking care of capitalization, elision and appropriate spacing around punctuation. It is also possible to wrap portions of text in HTML tags. The realizer accepts two types of input: constituents or dependents.

\subsubsection{Constituent notation} 
\label{ssub:constituent_notation}

In the constituent notation, the data structure is built by function calls whose names\footnote{Traditionally in \js{}, identifiers starting with a capital letter are constructors not functions. However in linguistics, symbols for constituents start with a capital letter so we kept this convention.} were chosen to be similar to the symbols typically used for constituent syntax trees\footnote{See the \href{http://rali.iro.umontreal.ca/JSrealB/current/documentation/user.html?lang=en}{documentation} for the complete list of functions and parameter types}:
\begin{itemize}
        \item \textbf{Terminal}: \lstinline!N! (Noun), \lstinline!V! (Verb), \lstinline!A! (adjective), \lstinline!D! (determiner) ...
        \item \textbf{Phrase}: \lstinline!S! (Sentence), \lstinline!NP! (Noun Phrase), \lstinline!VP! (Verb Phrase) ...
\end{itemize}
Features added to these structures using the dot notation can modify their properties. For terminals, their person, number, gender can be specified. For phrases, the sentence may be negated or set to a passive mode; a noun phrase can be pronominalized. Punctuation signs and  HTML tags can also be added. 

\subsubsection{Dependency Notation} 
\label{ssub:dependency_notation}
In the dependency notation, the data structure is built by function calls whose names are similar to the \emph{main} dependency relations as used in the Universal Dependencies~\cite{10.1162/coli_a_00402} (UD). UD relations were originally defined to be able to take into account most linguistic phenomena that occur in existing texts so many variations exist. But, in a generation context in French and English only, it was not felt necessary to keep all these nuances, so only a small subset of dependency relations was kept: 
\begin{itemize}
    \item \textbf{Dependent}: \lstinline!root!, \lstinline!subj!~(subject), \lstinline!comp!~(complement), \lstinline!mod!~(modifier), \lstinline!det!~(determiner) and \lstinline!coord!~(coordinate)
\end{itemize}
A dependency is specified by a function named after the dependency name taking as first parameter its head, a \textbf{Terminal} like in the constituent notation. The other parameters, if any, are other \lstinline!Dependent!s.

The organization of these relations follows roughly the conventions used in the Universal Dependencies~\cite{10.1162/coli_a_00402} except for \lstinline!coord! for which the head is taken as the coordination conjunction or a special symbol, perhaps an empty string;  the relations in dependents must all be the same (e.g. all \lstinline!subj! or \lstinline!comp!) which is considered as the global relation for the whole coordinate. In the \emph{traditional} UD notation, the head is instead the first coordinated term which is then linked to other members of the coordinate with a \lstinline!conj! relation. Given the fact that the number, gender and person of the dependents within a coordinate must be combined in order to compute the global number, person and number\footnote{This process is detailed in Section~\ref{coordination}}, it was found more practical to identify the coordinate with a special relation delimiting its dependents. 

Even though, we had initially chosen this representation of coordination for implementation purposes, it was then pointed out to us that Tesnière had proposed a similar view coordination which he referred to as \emph{jonction}~\cite[p 326]{Tesniere-59}.
\begin{quote}
La jonction s'opère entre deux noeuds de même nature, quelle que soit par ailleurs cette nature.[...] Mais il est indispensable que les deux noeuds soient de même nature.[...] on ne saurait joncter un actant et un circonstant, ni un noeud verbal et un noeud adjectival.\\  
\emph{[Junction works between two nodes of the same type, no matter this type. But it is necessary that the two nodes be of the same type, one cannot join an actant or a circonstant, neither a verbal node nor an adjectival node]}
\end{quote}

Realization sorts the dependents of a relation: \texttt{det} and \texttt{subj} are realized before the head, \texttt{comp} and \texttt{mod} are realized after the head. The ordering within the specification is used in the case of ties both before and after the head. A\lstinline! ("pre")! or \lstinline!("post")! feature can be specified to alter the default ordering to force its realization before or after the head. The position of adjectives is dealt according to the rules of English or French grammar, unless changed by \texttt{.pos}. 

\subsubsection{Programming aspects} 
\label{ssub:programming_aspects}
Since \jsr{} expressions are standard \js{} expressions built by functions and possibly modified with methods, no additional parsing of the input to \jsr{} is necessary. Usual browser-based \js{} development tools can thus  be used to develop the realizer functions while allowing a seamless integration in webpages. The special terminal \lstinline!Q! (\emph{Quote}) allows the insertion of \emph{canned text}, a feature that caters for special needs and is often useful in practical applications, but is not discussed further here. The Python implementation uses exactly the same notation; although it is less convenient within a web page because \py{} is not directly interpreted by the browser. On the other side, when used offline, it allows an easier integration with many other NLP tools.

\begin{figure}
{\small
\begin{lstlisting}
// CONSTITUENT notation
S(                     // Sentence
  Pro("him").c("nom"), // Pronoun (citation form), nominative case
  VP(V("eat"),         // Verb at present tense by default
     NP(D("a"),        // Noun Phrase, Determiner
        N("apple").n("p") // Noun plural
       ).tag("em")     // add an <em> tag around the NP
    )
 )
// DEPENDENCY notation
root(V("eat"),    // Sentence with a verb as head, two dependents
     subj(Pro("him").c("nom")), // Subject with a pronoun as head
     comp(N("apple").n("p"),    // Complement with a noun as head
          det(D("a")))          //    dependent with a determiner 
              .tag("em")       // add an <em> tag around the comp
    )
\end{lstlisting}
\hrule{}
\begin{lstlisting}[language=java,basicstyle=\footnotesize\ttfamily]
Lexicon lexicon = new XMLLexicon();     // default simplenlg lexicon
NLGFactory nlgFactory = new NLGFactory(lexicon);

NPPhraseSpec np= nlgFactory.createNounPhrase("a", "apple");// create NP
np.setFeature(Feature.NUMBER,NumberAgreement.PLURAL);   // set plural
SPhraseSpec s = nlgFactory.createClause("he","eat", np);// create sentence
DocumentElement sentence = nlgFactory.createSentence(s);

Realiser realiser = new Realiser(lexicon);
NLGElement realised = realiser.realise(sentence);
System.out.println(realised.getRealisation());
\end{lstlisting}}
\hrule
\begin{lstlisting}
from CoreNLG.DocumentConstructors import TextClass
class Content(TextClass):
    def __init__(self, section):
        super().__init__(section)
        self.text = self.free_text(
            "he",
            "eats",
            self.nlg_tags("em",text="apples")
        )
\end{lstlisting}
\caption{Top: \js{} using constituent or dependency notation with comments at the right realized by \jsr{} or \pyr{} as: \realization{He eats \textless{}em\textgreater{}apples\textless{}/em\textgreater{}.} Middle: \snlg{} Java statements to print a roughly equivalent sentence: \realization{He eats some apples.} Bottom: Python class from CoreNLG project to create a web page displaying \realization{He eats \emph{apples}}.}
\label{fig:input-notations}
\end{figure}

\jsr{} can also be used as a standalone \lstinline!node.js! module\footnote{Installable with \texttt{npm i jsreal}} taking input created by an external system. By experience, we know that creating syntactically valid \js{} expressions from a program written in another programming language can be cumbersome and tricky. Therefore \jsr{} also allows a JSON input format\footnote{The specifics are described in  \href{http://rali.iro.umontreal.ca/JSrealB/current/data/jsRealb-jsonInput.html}{this document}} for which convenient APIs are available in almost all programming languages. This idea is similar in principle to the XML input format allowed in \snlg{}. The JSON input format has been used to realize sentences from structures built by Prolog and Python programs. For Python, this is now less useful because of the port of \jsr{} to Python discussed in Section~\ref{sub:python_implementation}. 

The top part of Figure~\ref{fig:input-notations} shows our running example in both constituency and dependency notation, for the realization of the utterance \realization{He eats \textless{}em\textgreater{}apples\textless{}/em\textgreater{}.} in which the word \realization{apple} would appear emphasized when displayed in a browser.




\section{Previous text realizers} 
\label{sec:previous_work}

Realizers have been an integral part of Natural Language Generation (NLG) systems since the eighties. 
Realizers such as \kpml{}~\cite{KPML}, \surge{}~\cite{elhadad-robin-1996-overview},  \realpro{}~\cite{lavoie-rainbow-1997-fast}, Forge~\cite{mille-etal-2017-forge} and GenDR~\cite{lareau-etal-2018-gendr} are based on linguistic theories, taking into account many details in the construction of sentences, which allows powerful realizations. However, that complexity hinders their ease of use: writing specifications for them requires an intimate knowledge of the underlying theory. 

In fact, most existing realizers are considered so complex that \snlg{}~\cite{gatt-reiter-2009-simplenlg}, as its name implies, defines itself by its ease of learning and of use. Words, phrases and other structures are Java objects created and manipulated intuitively by a programmer and easily integrated into a Java project. While its principles somewhat limit the power of its realizations compared to other systems, these realizations are largely sufficient for many uses. \snlg{} has been used in a variety of text generation projects.\footnote{\snlg is freely available at \url{https://github.com/simplenlg/simplenlg}} Our realizer was originally inspired by \snlg{} for which we have developed a bilingual (French and English) version~\cite{Vaudry2013ENLG}.\footnote{Available at \url{https://github.com/rali-udem/SimpleNLG-EnFr}}

But with the rise of web applications, we found that the use of Java for building a text realizer was complicated by the need of creating a distinct web server for building new web pages and creating an XML structure for realization. As the Java notation is relatively verbose,  we decided to develop a notation that mimics the usual notation of the constituent grammar. Embedding the realizer in the web page itself greatly simplifies the architecture of the application. but \jsr{} would probably never have been created without our previous experience with \snlg{}. The middle part of Figure~\ref{fig:input-notations} shows the input to create a almost equivalent sentence to the one written in \js{}. \snlg{} does provide some HTML formatting but only at the paragraph level (list and title) and not at the token level; the plural of the indefinite article \realization{a} is realized as \realization{some}.

Many data-to-text information systems can \emph{get away} with template-based text generators in which it is sufficient to be able to deal with the description of a iist of items. Van Deemter et al. ~\cite{van-deemter-etal-2005-squibs} argue that these types of systems are interesting and useful and not necessarily \emph{inferior} to the more general approaches. 

One the first template system to be described in detail is YAG~\cite{Mcroy-2003} which bridges the gap between linguistic theories and the need for practical data-to-text generation systems.

\systeme{RosaeNLG}\footnote{\url{https://rosaenlg.org}} is a \js{} NLG library based on the Pug templating engine\footnote{\url{https://pugjs.org}}. It supports languages such as  English, French, German, Italian and Spanish with some features such as agreement within noun phrases, a few tenses and agreement at the third person of some simple tenses.  \systeme{RosaeNLG} was developed for realizing some simple data to text applications and is especially tuned for outputting lists of objects and properties using appropriate commas and a conjunction at end of the list. \systeme{RosaeNLG} being built over a template engine, it is not so well integrated within a web page as \jsr{} because a distinct specific file must be written and transformed into \js{} before being integrated in the web page.  Moreover the linguistic coverage of \jsr{}, for French and English at least, is much more complete than what is offered by \systeme{RosaeNLG}. 

\systeme{CoreNLG}\footnote{\url{https://github.com/societe-generale/core-nlg}} is a Python library for Natural Language Generation providing some tools to structure and write NLG projects, agreement between elements must be based on external resources. Like \jsr{}, it deals with some typographical conventions and elisions and also allows choosing randomly between synonyms, but it is mostly grammar unaware. A simplistic example of its formalism is shown in the bottom part of Figure~\ref{fig:input-notations}. 

\jsr{} differs from \systeme{compromise}\footnote{\url{http://compromise.cool}} which is a low-level \js{} API for both parsing English and extracting information.  It dubs itself as \emph{trying its best, and being small, quick, and usually good enough}. It also provides functions for generating English dealing with conjugation at simple tenses and with the plural of nouns. \systeme{compromise} stays at the level of the \js{} program and generates list of tokens, while \jsr{} is oriented toward the specification of the constituent structure from which the basic tools are automatically called. 

Text realizers are important in commercial systems such as ARRIA\footnote{\url{https://www.arria.com/}} that provide sophisticated templates that can be customized with a specialized programming language within well-developed web-based environments. The strength of the system lies with the ease of linking with different types of data sources and analytic tools, but it requires learning a specialized language instead of relying on the underlying and well-known \js{} formalism. 

Neural models for end-to-end data-to-text generation have been proposed, but they focus on the whole process and pay relatively little attention to the final realization step, which we address here. Castro Ferreira \emph{et al.}~\cite{castro-ferreira-etal-2019-neural} found that having intermediate steps in the generation process generally leads to better texts than end-to-end systems. This is one of the reasons why we devoted time to study this \emph{final} process of NLG.

We now detail some interesting aspects of the organization of \jsr{} to take care of the automatic agreement between constituents which is an important feature in order to limit the number of annotations to give and to add some flexibility when individual components are combined in different ways.


\section{Steps of the realization process}
\label{steps-in-the-realization-process}

\subsection{Structure creation}
\label{structure-creation}

\begin{figure}[ht]
  \centering
    \includegraphics[scale=\gscale]{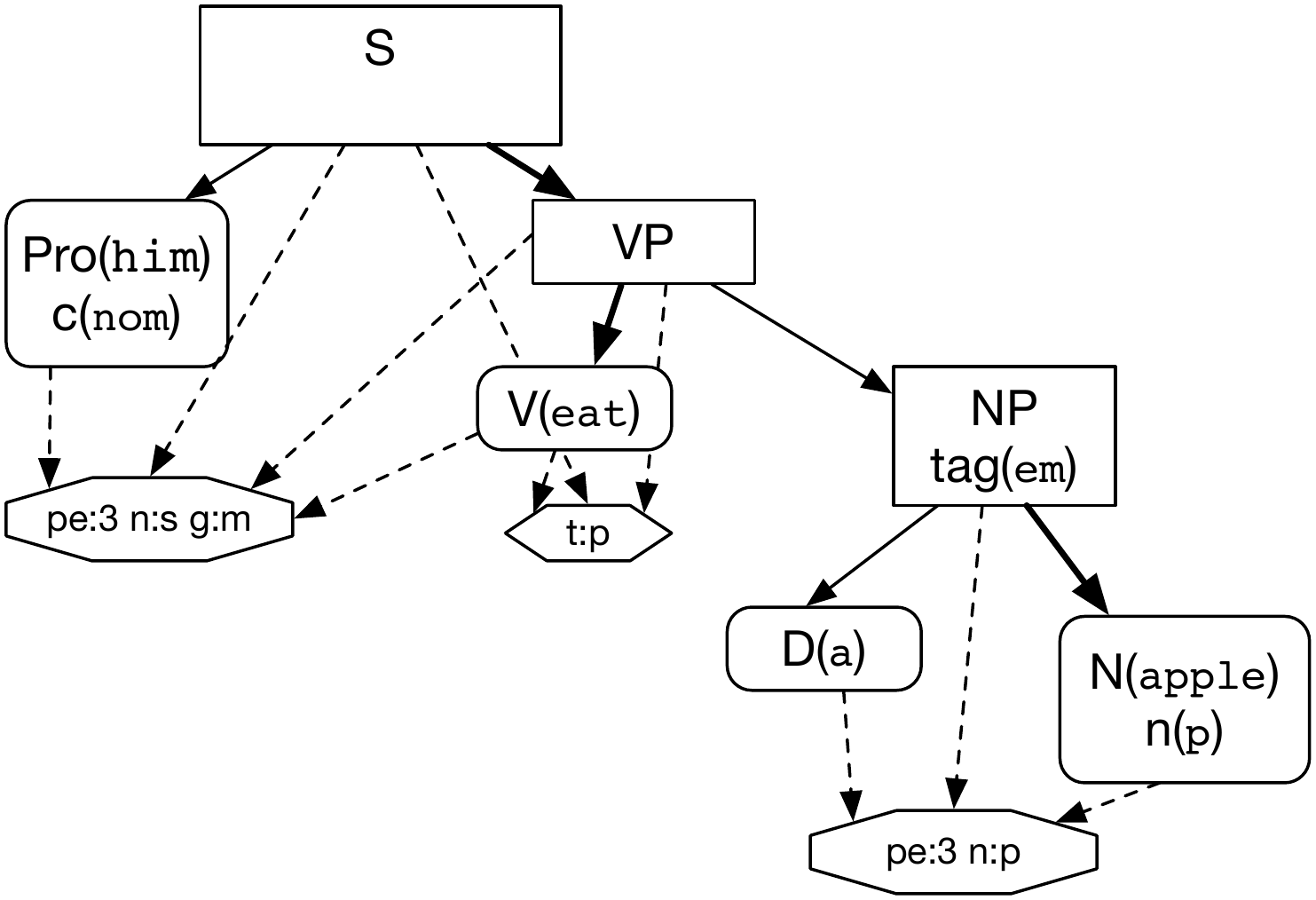}
  \caption{Internal data structure corresponding to the expression in constituent notation of Figure~\ref{fig:input-notations}}
  \label{fig:images_Active}
\end{figure}

To illustrate the steps to go from a \jsr{} expression, using the constituent notation, to the corresponding English sentence, we start with the sentence structure in Figure~\ref{fig:input-notations} whose corresponding internal data structure is shown in Figure~\ref{fig:images_Active}.

\begin{figure}[ht]
  \centering
    \includegraphics[scale=\gscale]{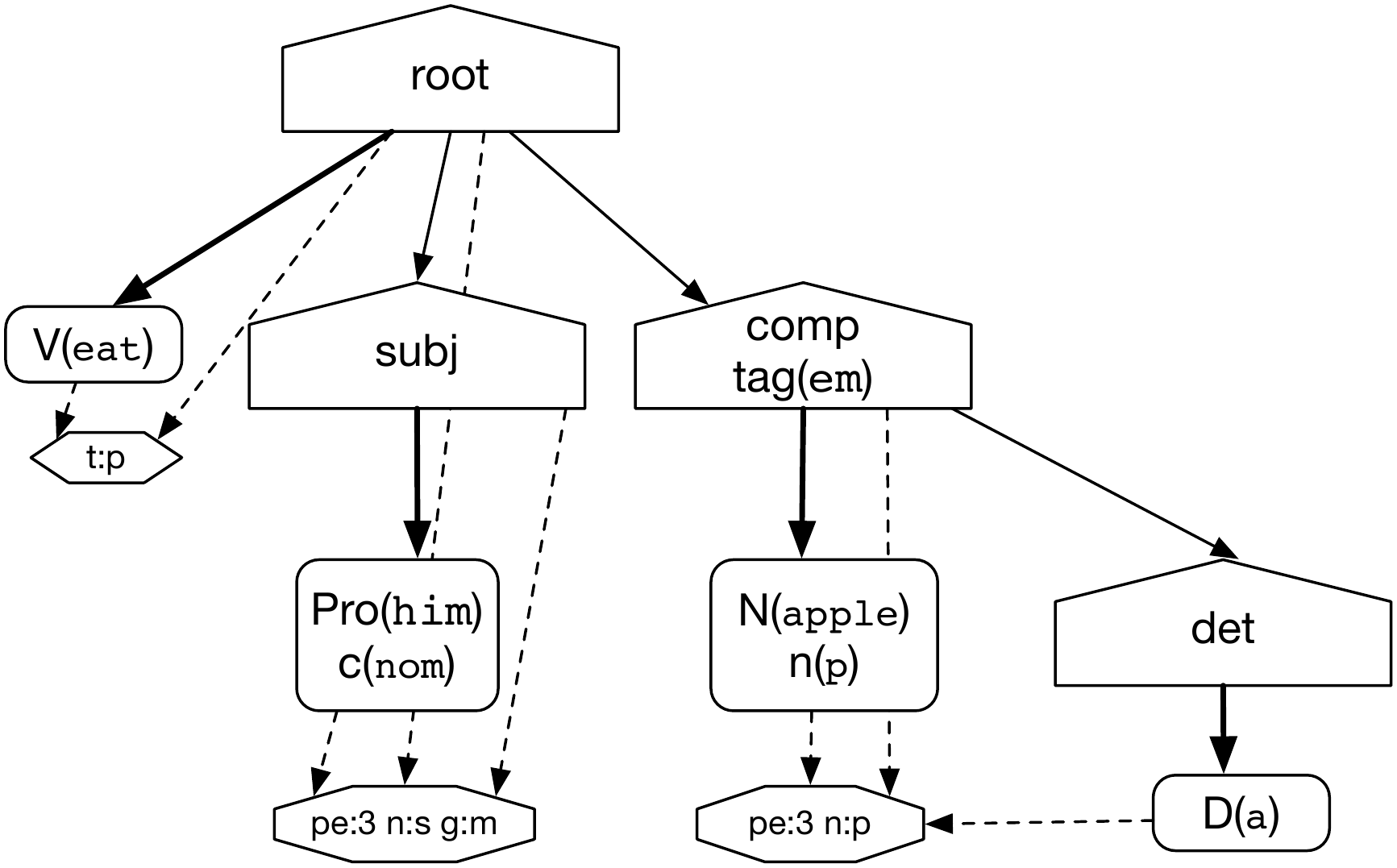}
  \caption{Internal data structure corresponding to the expression in dependent notation of Figure~\ref{fig:input-notations}}
  \label{fig:images_Active-dep}
\end{figure}

Diagrams in this paper use these graphical conventions:
\begin{itemize}
        \item a \textbf{Phrase} is shown as a rectangle containing its name on the first line and possibly options on the following line;
        \item a \textbf{Dependent} is shown as a pentagon (used in Figure~\ref{fig:images_Active-dep}) containing the name of the relation on the first line and possibly options on the next line;
        \item a \textbf{Terminal} (to be realized as a string of words, possibly none) is displayed as a rounded rectangle containing its name followed by its \emph{lemma} in parentheses; options may appear on the second line. On a third line, will be the realization string once it has been determined.
        \item Full line arrows indicate parent-child relationships, lines with larger width link a parent with its \emph{head} child. Dashed lines link a constituent with its \emph{shared} properties. Shared properties are first associated with terminals: they are person (\lstinline!pe!), number (\lstinline!n!) and gender (\lstinline!g!) (shown as a flat octagon in Figure~\ref{fig:images_Active}) for nouns, pronouns, adjectives, determiners and verbs; verbs also have two other shared properties, tense (\lstinline!t!) and auxiliary (\lstinline!aux!) (in French only), shown as a flat hexagon in Figure~\ref{fig:images_Active}. 
\end{itemize}


These properties are \emph{shared} to ensure that proper agreement is done between constituents\footnote{In Prolog, this would be achieved by unification between shared variables that are changed as soon as one of its occurrences is changed.} so that when a property of constituent is changed then all dependent constituents will be changed accordingly. 

When phrases are created from terminals and other phrases, the shared properties of the parent are set to the ones of its head so that modifications to the parent will be propagated to its children. In Figure~\ref{fig:images_Active}, the tense of the \lstinline!VP! is set to the one of the \lstinline!V!, and the number of the \lstinline!NP! is set to the one of the \lstinline!N!.  But other links must also be added: a \lstinline!VP! must also be linked with its subject in order to properly agree in person, gender and number, this is why the \lstinline!VP! is linked with the shared property of the \lstinline!Pro!. Agreement must also be made between children of a \lstinline!NP!: a determiner, adjectives must agree with the head of a \lstinline!NP!, so they all share the same properties.

Some properties (such as formatting) are not shared, they only apply to the current constituent. The only shared properties are those dealing with an agreement between constituents.

Figure~\ref{fig:images_Active-dep} shows the diagram of the internal data structure when using the dependent notation with the agreement links. One interesting implementation advantage of this notation is the fact that the head of a relation is explicitly given and does not have to be found using heuristics described previously. Once the appropriate agreement links are set up, propagation is performed as for the constituent notation.

\subsection{Realization of Constituents} 
\label{sub:realization_of_constituents}

We now explain how to go from an input structure to an English sentence. To realize a sentence, the \lstinline!toString()! function performs two tasks:
\begin{itemize}
        \item \emph{stringification} which computes the realization property of a terminal possibly modified with some formatting;
        \item \emph{detokenization} which creates a single final string for a phrase.
\end{itemize}

These steps, which are the same for both the constituent and the dependency notations, are now detailed.

\subsubsection{Stringification of constituents}
\label{stringification-of-constituents}

The production of strings from terminals and phrases is carried out recursively. Realization strings are produced for each terminal, but also for each phrase at each level of the tree. This process builds lists containing the original terminals, using information from the original data structure (e.g.,~number, gender or agreement links) for proper conjugation and declension.

\begin{figure}[ht]
  \centering
    \includegraphics[scale=\gscale]{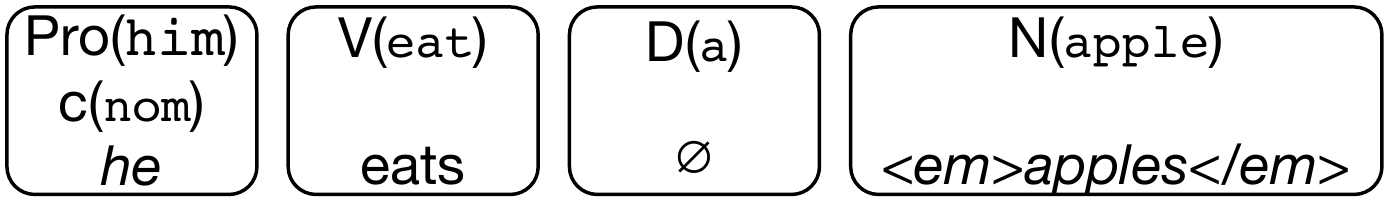}
  \caption{Tokens from Figure~\ref{fig:images_Active} once their \lstinline!realization! field is filled in.}
  \label{fig:images_Active-Realization}
\end{figure}
\subsubsection{Computing the \texttt{realization} property}
\label{computing-the-realization-property}

The \lstinline!toString()! function of a \emph{Phrase} calls \lstinline!toString()! on its children to create into a flat list containing the original terminals with their realization field filled in.

The \lstinline!toString()! function of a \emph{Terminal} applies declension or conjugation rules taking into account the gender, number, person and possibly tense information in order to fill in the \lstinline!realization! field of the terminal. Figure~\ref{fig:images_Active-Realization} illustrates a few interesting cases.

\begin{itemize}
\item \lstinline!Pro("him").c("nom")! is realized as \realization{he} because the declension rules set its gender to masculine, its person to third and its number to singular;
\item \lstinline!VP("eat")! is realized by the string \realization{eats} because the verb is conjugated at the present tense by default as given by its shared property and, as shown in Figure~\ref{fig:images_Active} it is as third person singular, from the shared property of its subject, the \lstinline!Pro!;
\item \lstinline!D("a")! produces an empty string because it is a plural indefinite article as given by its shared property with the \lstinline!N!.
\item \lstinline!N("apple").n("p")! produces \realization{\textless{}em\textgreater{}apples\textless{}/em\textgreater{}}, the plural declension of the lemma \lstinline!apple!. As will be explained in the next section, the formatting \realization{em} tags come from the surrounding \lstinline!NP!.
\end{itemize}

Using the shared properties allows global modifications to the original expression, although it does not override locally set properties. For example, adding \lstinline!.t("ps")! at the end of the expression in Figure~\ref{fig:input-notations} indicates that the whole expression should be at the simple past. The modified expression would then be realized as \realization{He ate \textless{}em\textgreater{}apples\textless{}/em\textgreater{}.}

\subsubsection{Formatting the realization property}
\label{formatting-the-realization-property}

Formatting is carried in a series of optional steps:
\begin{enumerate}
\item Apply language-dependent token modification rules taking into account the surrounding words:
  \begin{itemize}
  \item English:
    \begin{itemize}
    \item the determiner \lstinline!a! which changes to \lstinline!an! depending on the first letter of the next word, most often a vowel.
    \end{itemize}
  \item French:
    \begin{itemize}
    \item \emph{elision} for determiners such as \lstinline!le!, \lstinline!ce!, \lstinline!la!, etc., or conjunctions such as \lstinline!que!,
\lstinline!puisque!. If the next word starts with a vowel, then the current word's last letter is changed to an apostrophe \lstinline!'!, which will be joined with the next word. In fact, rules for French are a little more complicated because this case also applies to some words starting with an \lstinline!h!. 
    \item \emph{euphony} for adjectives such as \lstinline!beau!, \lstinline!fou! or \lstinline!vieux! which are changed to \lstinline!bel!, \lstinline!fol! or \lstinline!vieil! if the next word starts with a vowel.
    \item \emph{contraction} combining words such as \lstinline!de le! to \lstinline!du!, \lstinline!si il! to \lstinline!s'il!, \lstinline!à les! to \lstinline!aux!.
    \end{itemize}
  \end{itemize}
\item Modify the realization string to insert strings that should appear before \lstinline!.b(..)!, after \lstinline!.a(...)! or around \lstinline!.ba(...)! the realization string;
\item Surround the realization string with HTML tags given by \lstinline!.tag(...)!.
\end{enumerate}

In our example, only the HTML tag \lstinline!em! was added when formatting the \lstinline!NP!. Section~\ref{formatting} will discuss why formatting can interfere with the realization process.

\subsubsection{Detokenization}
\label{detokenization}

When the \lstinline!toString()! method is called for a constituent at the \emph{top-level} of an expression, a \emph{detokenization} process is applied to produce a well-formed English sentence. In our running example, this produces \realization{He\ eats\ \textless{}em\textgreater{}apples\textless{}/em\textgreater{}.}

Detokenization creates a single string by inserting appropriate whitespace between the realization strings of the terminals. When detokenization is applied to the top-level \lstinline!S!, it capitalizes the first letter of the sentence and adds a full stop at the end unless indicated otherwise.

We thus see that even for our example's simple \emph{three words} sentence, many steps are needed to produce a well-formed sentence.

\begin{figure}[ht]
  \centering
    \includegraphics[width=.9\textwidth]{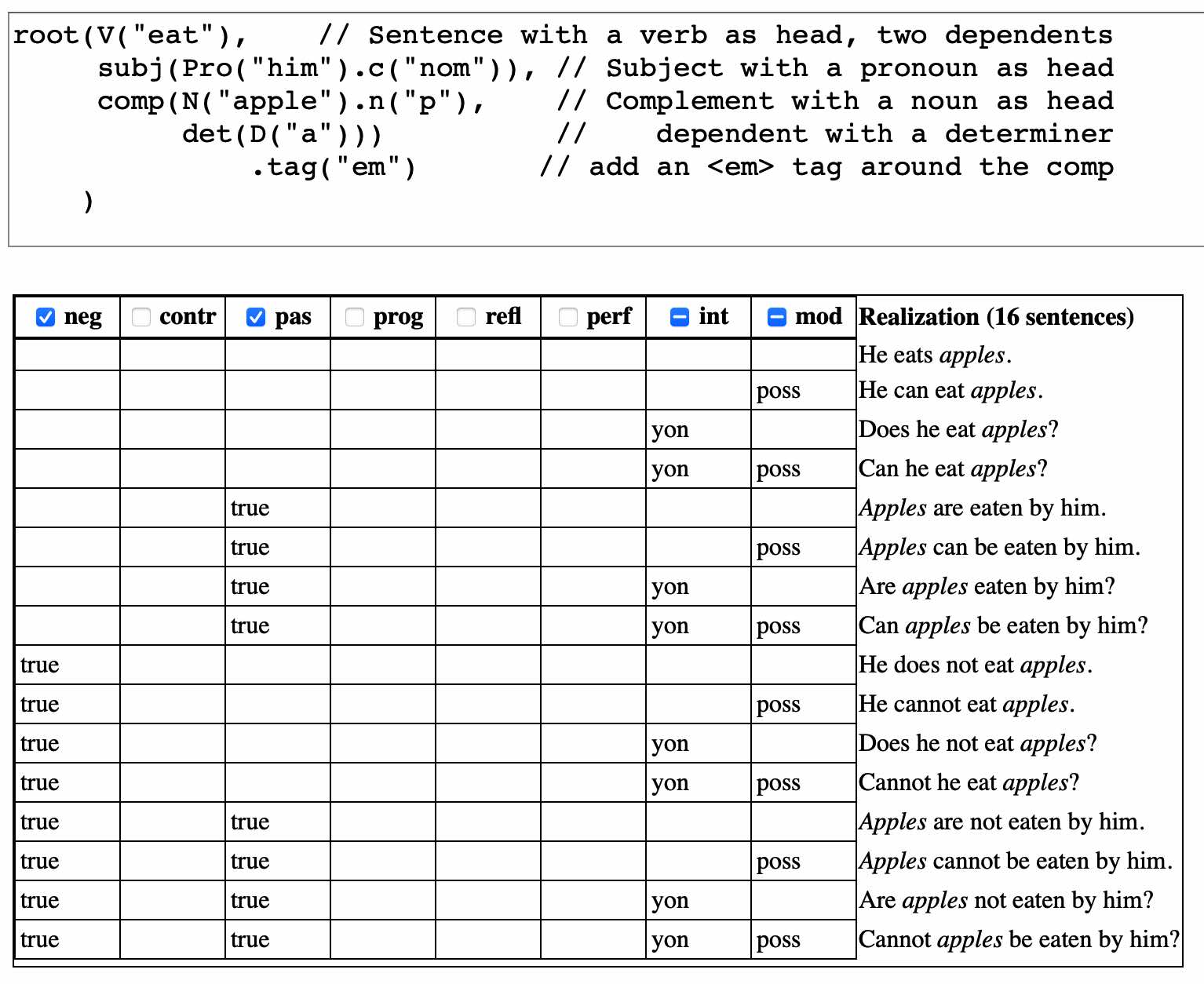}
  \caption{Variations on the sentences of Figure~\ref{fig:input-notations} combining negation, passive, yes or no question and possibility modality.}
  \label{fig:SentenceTypes}
\end{figure}

\section{Structure modifications}
\label{structure-modifications}

One very useful feature of \jsr{}, inspired by a similar one in \snlg{}, is the fact that a single affirmative sentence structure can be realized as negative, passive, interrogative or with a modality verb by setting a \emph{type} flag on the sentence or verb phrase structure without having to change its structure. Almost 5~000 variations can be generated from a single English affirmative sentence and 1~250 variations for a French\footnote{See this  
\href{http://rali.iro.umontreal.ca/JSrealB/current/demos/VariantesDePhrases/index.html}{demonstration of these variations with example sentences}}, even though some of them are not very colloquial or would need a specific context to be valid, Figure~\ref{fig:SentenceTypes} shows a few of them.
This capability of the realizer proved very useful when verbalizing AMR structures in which negation is only indicated by a negative polarity role on an otherwise affirmative structure. We could just translate the rest of the sentence as an affirmative one and add a negative flag at the end. This was also used for creating negations for augmenting a corpus of negative sentences for training a neural semantic analyzer.

To realize these modifications, \jsr{} modifies the original sentence structure by adding new words and changing word order while keeping agreement links intact. This section highlights a few interesting cases using the constituent notation, but the process is similar to the dependent notation.
\clearpage
\subsection{Negation}
\begin{figure}[ht]
{\small
\begin{lstlisting}
        S(Pro("him").c("nom"),
          VP(V("eat"),
             NP(D("a"),N("apple").n("p")))
         ).typ({neg:true})
\end{lstlisting}}
  \centering
    \includegraphics[scale=\gscale]{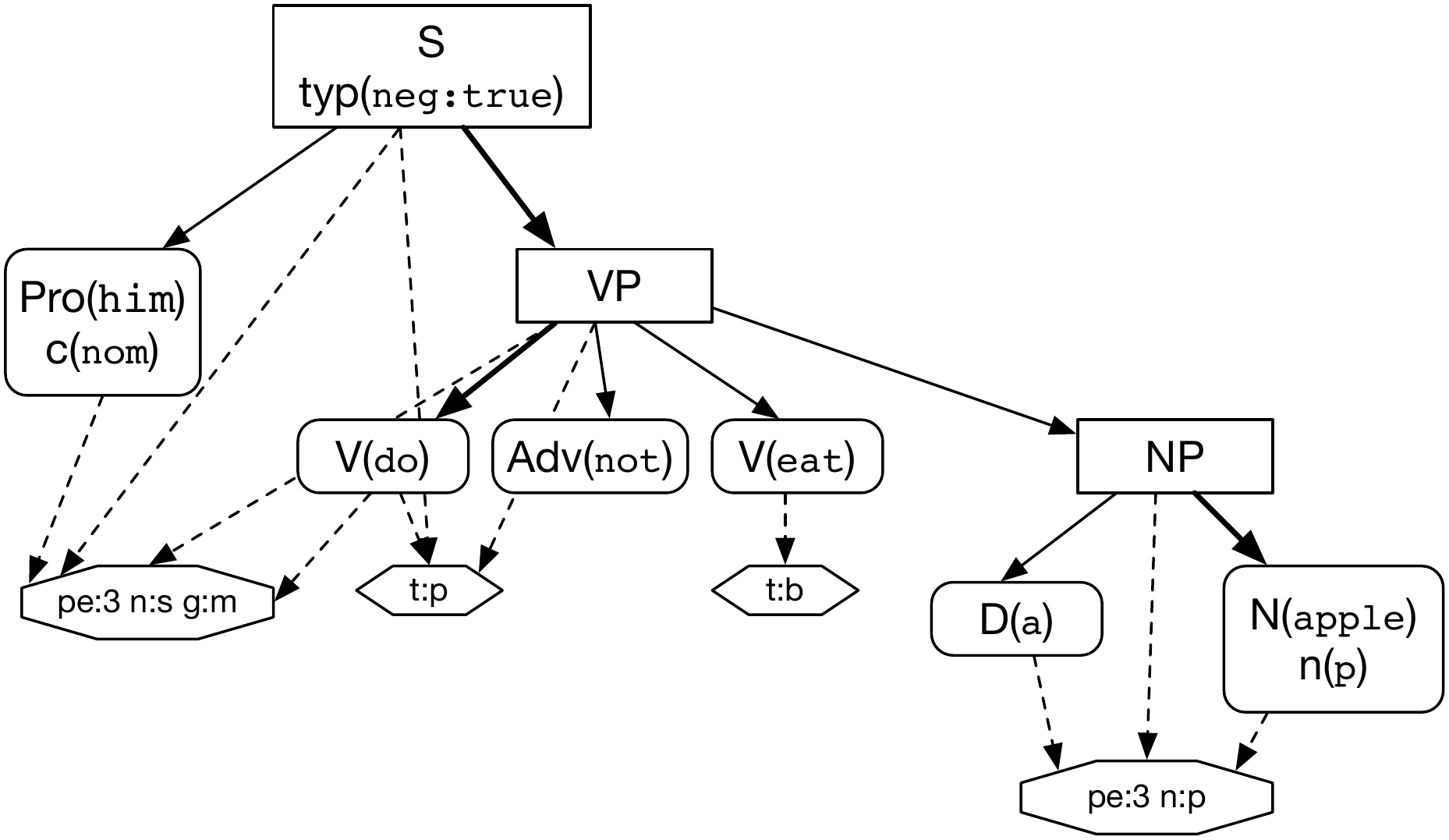}
    
    \includegraphics[scale=\gscale]{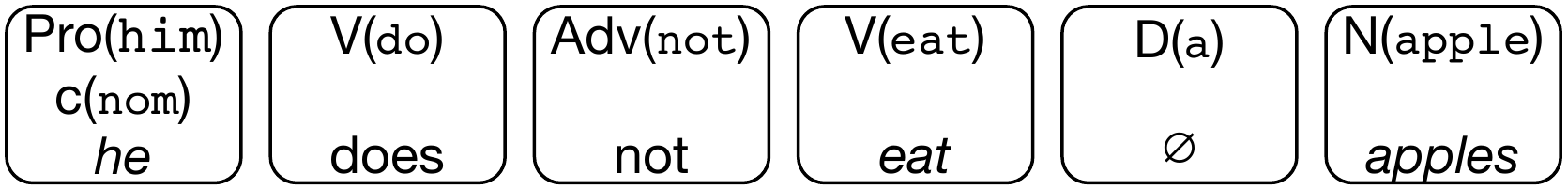}
  \caption{At the top is the example of Figure~\ref{fig:images_Active} with a negation flag set; in the middle is shown the modified data structure produced by \jsr{}; the bottom shows the resulting list of tokens.}
  \label{fig:images_Negative}
\end{figure}
We now explain how a negative sentence is realized when \lstinline!.typ({neg:true)}! is added to the top-level \lstinline!S! constructor of our previous example. To simplify the diagram, we removed the \lstinline!.tag("em")! from the NP. Figure~\ref{fig:images_Negative} shows the \js{} notation (top) and the resulting data structure (middle) once the \lstinline!.typ(..)! is applied. The structure of the \lstinline!VP! is modified. The auxiliary \realization{do} and the adverb \realization{not} are inserted in front of the verb. 
The head of the VP is now the auxiliary \lstinline!do! which shares its property with the \lstinline!Pro!. The tense of the original verb is now infinitive (\lstinline!b! for \emph{base form}).
The \emph{stringification} and \emph{detokenization} processes are the same as in the previous example. The end result is the string \realization{He does not eat apples.}
\clearpage

\subsection{Passivization}
\label{passivization}
\begin{figure}[ht]
{\small
\begin{lstlisting}
        S(Pro("him").c("nom"),
           VP(V("eat"),
              NP(D("a"),N("apple")).n("p"))
         ).typ({neg:true,pas:true})
\end{lstlisting}}
  \centering
    \includegraphics[scale=\gscale]{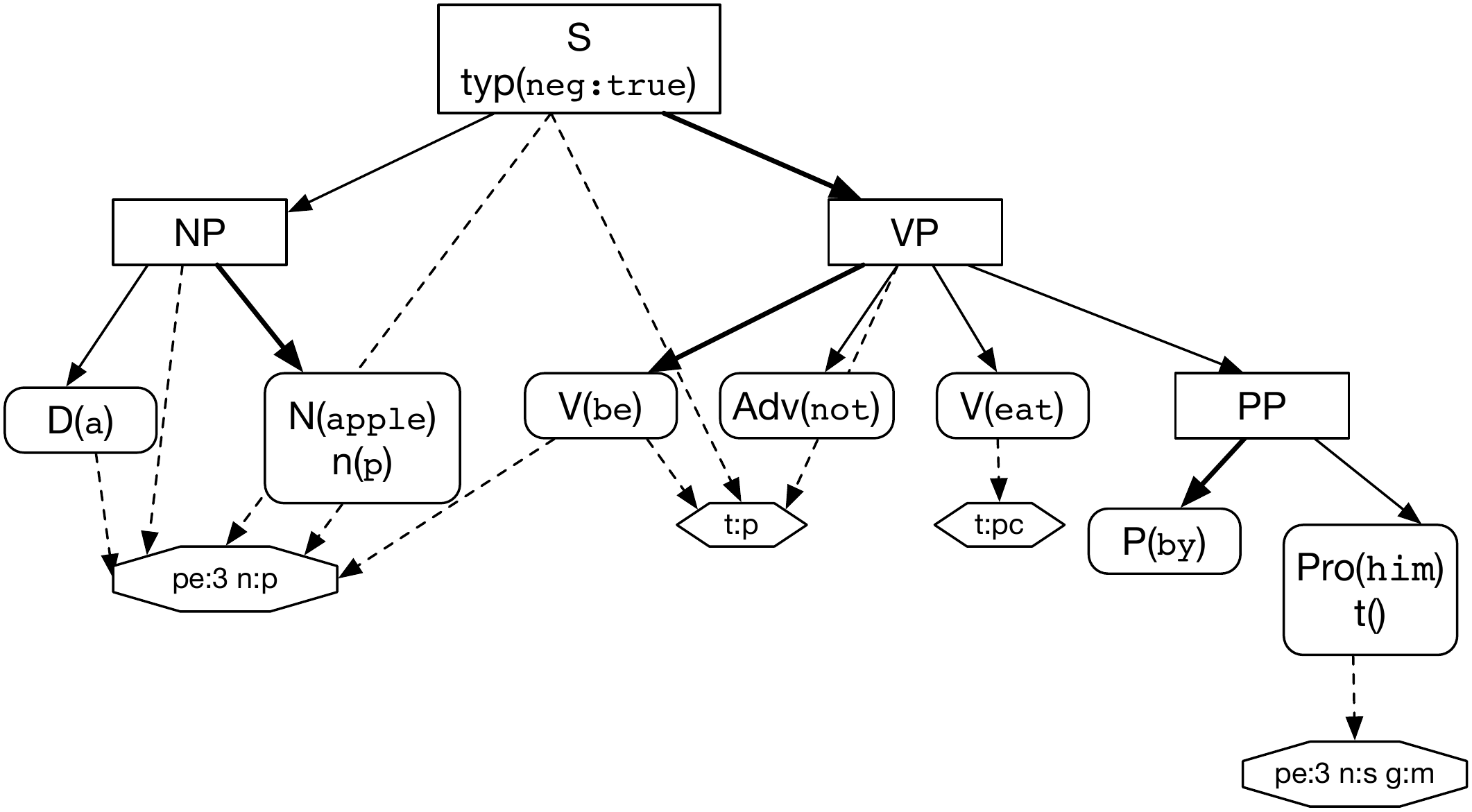}
    
    \includegraphics[scale=\gscale]{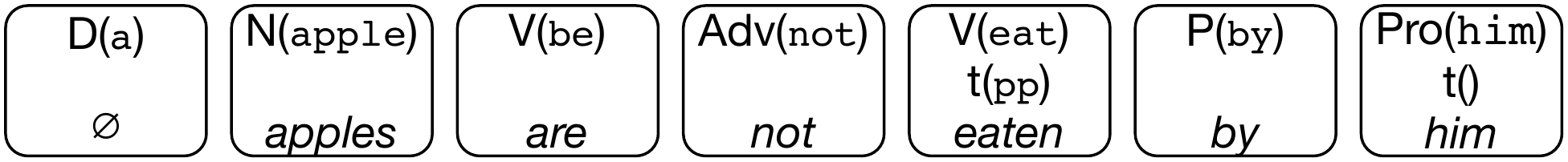}
  \caption{Figure~\ref{fig:images_Negative} with passive flag set and the resulting data structure and tokens after passivization has been performed. The final string is \realization{Apples are not eaten by him.}}
  \label{fig:images_Negative-Passive}
\end{figure}

Figure~\ref{fig:images_Negative-Passive} shows the transformation to the \emph{passive} form of our running example. It is now realized as \realization{Apples are not eaten by him.} In the \js{} notation at the top, \lstinline!pas:true! is added in the \lstinline!.typ! call. The middle part of the figure shows the data structure once these negative and passive transformation processes are applied: the object becomes the subject, the main verb becomes \lstinline!be!, the original verb's tense is changed to the past participle, and the original subject becomes a prepositional phrase starting with \lstinline!by!. In our example, the original subject being a nominative pronoun, it has to be changed to its tonic form. 
The head of the \lstinline!VP! is now \lstinline!be! whose shared properties for the subject is the \lstinline!NP!
As the subject changes, the shared link for the \lstinline!S! must also be changed. 

\subsection{Question generation} 
\label{sub:question_generation}

\begin{figure}[ht]
  \centering
    \includegraphics[width=.45\textwidth]{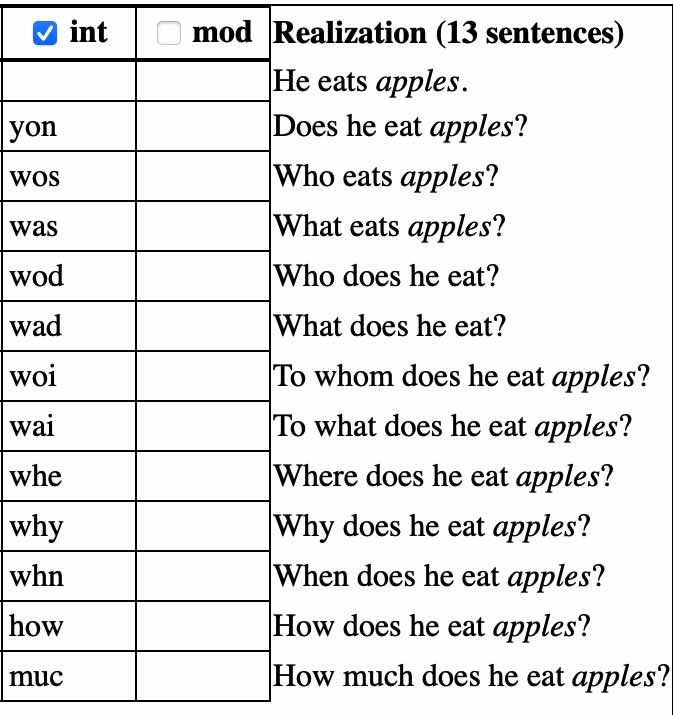}
  \caption{Twelve types of questions that can be generated from a single affirmative sentence}.
  \label{fig:questions}
\end{figure}

\begin{figure}[hb]
  \centering
    \includegraphics[width=0.3\textwidth]{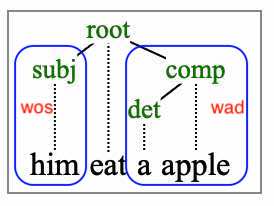}
  \caption{Dependency tree of Figure~\ref{fig:input-notations} showing in blue the parts that can be removed to create two questions: \lstinline!wos! (who-subject) \realization{Who eats an apple ?} and \lstinline!wad! (what-direct object) \realization{What does he eat?}}
  \label{fig:DepTree}
\end{figure}

Creating questions from an affirmative sentence is quite productive as shown in Figure~\ref{fig:questions}. In order to create questions from a single dependency structure, jsRealB uses the \emph{classical} grammar transformations: for a \emph{who} question, it removes the subject (i.e., the \lstinline!subj! dependent), for a \emph{what} question, it removes the direct object (i.e., the first \lstinline!comp! dependent), for other types of questions (\emph{when},\emph{where}) it removes the a \lstinline!mod! or \lstinline!comp! having a preposition as its head. Figure~\ref{fig:DepTree} illustrates this process on our running example.

Creation of questions from parsed affirmative sentences was used to build a training corpus for a neural question-answering system~\cite{LeBerre2021EMNLP}. Almost 200,000 questions were generated from a corpus of about 140~000 sentences.
In order to determine which questions are appropriate for a given sentence, the dependency structure of the original sentence is checked to determine if it contains the required part to be removed. Depending on the preposition, the question will be a \emph{when}, a \emph{where} or another type of question.  An interesting side product is the fact that the \emph{removed} part is the answer to the generated question, which is very useful in a teaching context or for generating distractors in multiple-choice questions. Pronoun only answers (such as \lstinline!he! for the \lstinline!wos! (who as subject) question in Figure~\ref{fig:DepTree}) are not usually informative, so they were ignored in building the training corpus.


\subsection{Pronominalization}
\label{pronominalization}
\begin{figure}[ht] 
{\small
\begin{lstlisting}
        S(Pro("lui").c("nom"),
          VP(V("donner").t("pc"),
             NP(D("un"),N("pomme")).pro()
         )
\end{lstlisting}}
  \centering
    \includegraphics[scale=\gscale]{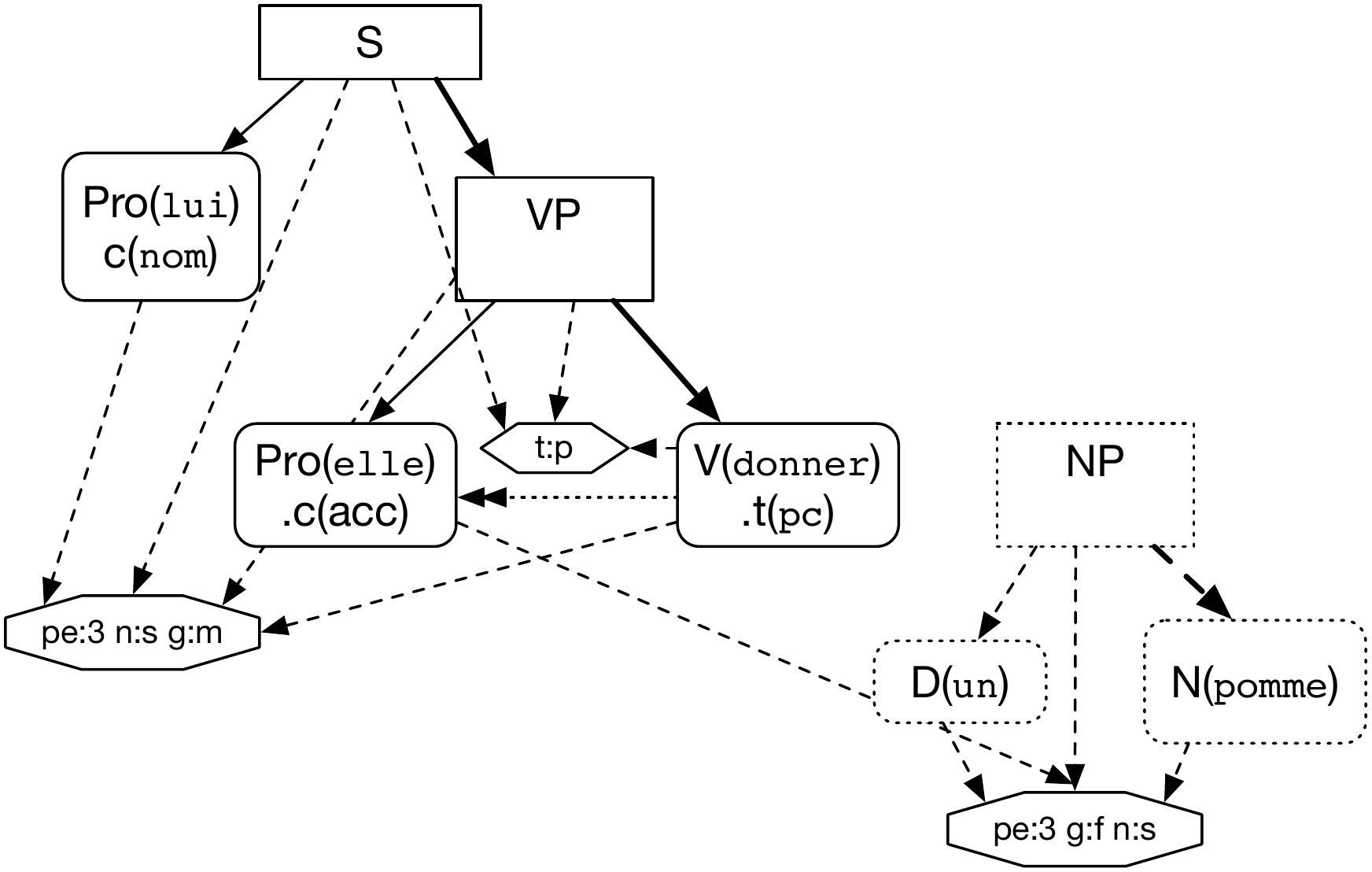}
    
    \includegraphics[scale=\gscale]{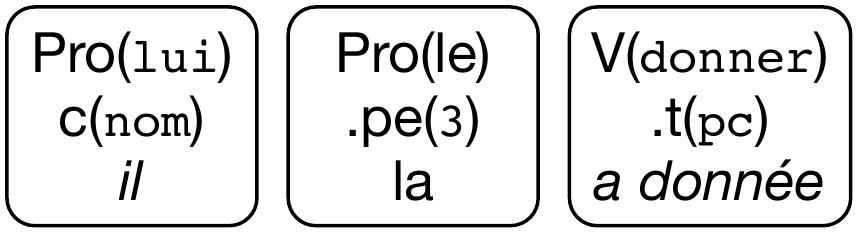}
  \caption{\js{} specification of a French sentence and the resulting data structure after pronominalization}
  \label{fig:images_French-Active-Pronoun}
\end{figure}

Another important structure transformation is the pronominalization process to change a noun or prepositional phrase to a pronoun having the same number and gender as the original phrase. That  process is indicated by calling \lstinline!.pro()! method on the phrase. We use an example in French because it illustrates more intricate transformation processes\footnote{In English, this would be realized simply as \realization{He has given it.}} coupled with agreement difficulties within the realization process. 

Not taking into account the call to \lstinline!.pro()!, the top of Figure~\ref{fig:images_French-Active-Pronoun} shows a variation applied to the French version of our running example. It is realized as \realization{Il a donné une pomme.} (\emph{He has given an apple.}) in which the verb is conjugated to \emph{passé composé}, corresponding approximately to the English \emph{present perfect} in its form. When pronominalization is applied, the realized sentence is \realization{Il l'a donnée.} (\emph{*He it has given.})

Figure~\ref{fig:images_French-Active-Pronoun} shows the data structure once the call to \lstinline!.pro()! has been processed: the noun used as the direct object must be replaced by an accusative pronoun. In this case \lstinline!Pro("elle").c("acc")! agreeing in gender (feminine) and number (singular) with the original noun that will not be realized. This is why it is shown as a dashed rounded rectangle in the figure.

In French, some \emph{interesting} peculiarities must be taken into account:
\begin{itemize}
\item a pronoun used as the direct object must appear \textbf{before} the verb (except for an imperative verb), so the order of the children of the \lstinline!VP! it must be changed 
while linking the pronoun with the properties of the original noun to ensure proper gender and number;
\item A verb conjugated at the \emph{passé composé} is built using an auxiliary verb (\realization{avoir} (\emph{have}) in this case) followed by a past participle;
\item a past participle with the \realization{avoir} auxiliary must agree in gender and number with its direct object when it appears before in the sentence. This was not the case in the original sentence, but once the pronoun is shifted before the verb, the past participle must agree with the pronoun. 
This is why in French a supplementary link (shown here as a dotted line with a double arrow head) between a verb and its direct object. So when the past participle is realized, it checks its position and then uses the gender and number of the direct object, here the \lstinline!Pro(elle)! whose shared properties are the ones of the original \lstinline!NP!. But the auxiliary of the main verb must agree with the original subject.
Another case of past participle agreement occurs with the \realization{être} (\emph{be}) auxiliary must agree with its subject. We will see later an instance of this case.
\end{itemize}\emph{}

The modified structure is detokenized as in the preceding section to produce the tokens shown at the bottom of Figure~\ref{fig:images_French-Active-Pronoun}.  Elision transforms \realization{la a} to \realization{l'a} and the formatting is realized as: \realization{Il l'a donnée.}

\begin{table}[h]
\begin{tabular}{|>{\vspace*{1ex}}p{.05in}|p{3.45in}|p{2.5in}|}
\hline
1 &\begin{lstlisting}
S(Pro("lui").c("nom"),
  VP(V("donner").t("pc"),
     NP(D("un"),N("pomme")).pro())
 )@\underline{.typ({neg:true}}!)
\end{lstlisting}&
{\vspace*{1ex}Il \underline{ne} l'a \underline{pas} donnée.\newline
\emph{*He it did \underline{not} give.}}
\\
\hline
2&\begin{lstlisting}
S(Pro("lui").c("nom"),
  VP(V("donner").t("pc"),
     NP(D("un"),N("pomme"))@\underline{.pro()}!,
     @\underline{PP(P("à"),
             NP(D("le"),N("fille")))}!)
 ).typ({neg:true})
\end{lstlisting}&
{\vspace*{1ex}Il ne l'a pas donnée \underline{à la fille}.\newline
\emph{*He it did not give \underline{to the girl}.}}
\\
\hline
3&\begin{lstlisting}
S(Pro("lui").c("nom"),
  VP(V("donner").t("pc"),
     NP(D("un"),N("pomme"))@\underline{.pro()}!,
     PP(P("à"),
        NP(D("le"),N("fille")))
           @\underline{.pro()}!)
 ).typ({neg:true})\end{lstlisting}&
{\vspace*{1ex}Il ne la \underline{lui} a pas donnée.\newline
\emph{*He \underline{it} to her did not give.}}
\\
\hline
4&\begin{lstlisting}
S(Pro("lui").c("nom"),
  VP(V("donner").t("pc"),
     NP(D("un"),N("pomme"))@\underline{.pro()}!,
     PP(P("à"),
        NP(D("le"),N("fille")))
           .pro())
 ).typ({neg:true,@\underline{pas:true}!})\end{lstlisting}&
{\vspace*{1ex}Elle ne lui \underline{a} pas \underline{été donnée} \underline{par lui}.\newline
\emph{*It \underline{has} not \underline{been given} \underline{to her} \underline{by him}.}}
\\
\hline
\end{tabular}
  \caption{\js{} specifications and the corresponding realizations for combinations of pronominalization, negation and passivation as produced by \jsr{}.}
  \label{tab:modifications}
\end{table}
\subsubsection{Further modifications}
\label{further-modifications}

French negation combined with pronominalization raise delicate word ordering problems. Table~\ref{tab:modifications} illustrates modifications of the example of Figure~\ref{fig:images_French-Active-Pronoun} in which the differences between each example are underlined. The second column gives the \jsr{} specification and the third column shows the corresponding French realization with an English transliteration in italics. 
\begin{itemize}
\item In line 1, \realization{ne\ ...\ pas} wraps around the verb and the preceding pronoun.
\item Line 2 adds an indirect object using a \lstinline!PP! (prepositional phrase) at the end of the sentence.
\item When the \lstinline!PP! is pronominalized (line 3), the  pronoun must also appear before the verb; as the pronoun is combined with \realization{à}, then the dative form \realization{lui} must be used.
\item Line 4 shows the passive form in which \realization{par lui} \emph{(by him)} comes from the tonic form of the pronoun that was the subject. In this case, the past participle is still singular feminine but for a different reason. Passivization takes the direct object and makes it the subject of the passive sentence used with the \realization{be} auxiliary. So the past participle must agree with the new subject which, in this case, was the original object occurring before the verb. 
\end{itemize}
\clearpage
These examples might seem contrived, but these cases appear in real world texts and it is important that they be handled correctly to create correct French texts. \jsr{} implements the French clitic placement rules as defined by the \emph{Encyclopédie grammaticale du français}~\cite{EGF-clitiques}. They also show the importance of having a distinct realization step that can take care of these cases and make it possible to realize many variations from a single input structure. 

English passivization also raises interesting problems especially when combined with perfect tenses, progressive mood and modal verbs (possibility, permission, necessity ...) for which \emph{affix hopping} rules~\cite[pp 38--48]{SyntStruct02Chomsky} are implemented. 

\section{Other useful features to take into account}
\label{other-interesting-features-to-take-into-account}

\subsection{Incremental building of \emph{phrases}}
\label{incremental-building-of-phrases}

Although our previous examples of \jsr{} expressions have been created manually, these expressions are most often built by programs by invoking the API's \js{} functions. Moreover, it may happen that not all arguments to a phrase are known before calling the function. For example, the subject and the verb can be determined in one part of a program, but complements are only later specified. To account for this possibility, \jsr{} allows adding new elements to an existing phrase. The \lstinline!add(Constituent,position)! method inserts either a phrase or a terminal to the current phrase at a certain position given by a non-negative index. If \lstinline!position! is not specified, the constituent is added at the end. 

For example
\begin{lstlisting}
        S(Pro("him").c("nom"),
          VP(V("eat"),
             NP(D("a"),N("apple").n("p")).add(A("red")))
         ).add(Adv("now").a(","),0)
\end{lstlisting}
\noindent
is realized as \realization{Now, he eats red apples.}. The adjective \lstinline!red! is added at the end of the \lstinline!NP! but, because adjectives in English are placed before the noun, it appears before the noun. The adverb \realization{now} followed by comma is inserted at the start of the sentence because the \lstinline!position! is set to 0. 

This dynamic feature explains why most realization decisions are taken at the very last moment (i.e., when a string is needed) and not while the structure is being built. 

\subsection{Coordination}
\label{coordination}

Data to text applications must often output lists of objects that can be realized using template-based systems like CoreNLG and RosaeNLG described in Section~\ref{sec:previous_work}. But linguistically, these descriptions use 
coordinated phrases which are specified in \jsr{} with a phrase \lstinline!CP! constituent or a \lstinline!coord! dependency in which a conjunction is specified, if any, with as many elements or dependents as needed. In the corresponding realization, all elements except for the last one are separated by a comma followed by the conjunction and the final element. The following example (shown in both constituency and dependency notations)

\begin{lstlisting}
S(CP(C("and"),NP(D("the"),N("apple")),    
              NP(D("the"),N("orange")),   
              NP(D("the"),N("banana"))),  
  VP(V("be"),A("good")))                  
//=====================================                                          
root(V("be"),   
     coord(C("and"),
           subj(N("apple"),det(D("the"))),
           subj(N("orange"),det(D("the"))),
           subj(N("banana"),det(D("the")))),
     mod(A("good")))  
\end{lstlisting}
\noindent
is realized as \realization{The\ apple,\ the\ orange\ and\ the\ banana\ are\ good.} in which \jsr{} takes into account that the subject is now plural because of the \realization{and}. If \lstinline!C("or")! had been given, the verb would have been realized as singular.

Such coordinated sentences are often built incrementally and, in some cases, only one element is required. For example,
\begin{lstlisting}
S(CP(C("and"),NP(D("the"),N("apple")),    
  VP(V("be"),A("good")))                  
//====================================
root(V("be"),
     coord(C("and"),
           subj(N("apple"),det(D("the")))),
     mod(A("good")))                                          
\end{lstlisting}
\noindent
is realized as \realization{The apple is good.} in hich the conjunction is ignored and the number stays singular, unless, of course, the single subject is plural.

The generation of the tokens for a coordinated phrase must also be performed at the \emph{last minute}, i.e.,~during stringification, where this special case must be checked.

\subsection{Reusing \jsr{} expressions}
\label{reusing-jsrealb-expressions}

One of the advantages of using a programming language for creating text is the fact that repetitive structures can be coded once and reused as often as needed. \jsr{} expressions, being \js{} objects, can be saved in variables, received as parameters or returned as result by functions. This fact is heavily used in the \href{http://rali.iro.umontreal.ca/JSrealB/current/Tutorial/tutorial.html}{demos and tutorial}.

As we have shown earlier, options may modify the original structure of the expression. It is worth pointing out that reusing a modified expression will thus realize the modified object and not the original one according to the usual \js{} behavior. For example, given the assignment:

\begin{lstlisting}
        let apple = NP(D("a"),N("apple"))
\end{lstlisting}
\noindent
the expression
\begin{lstlisting}
        S(Pro("him").c("nom"),
          CP(C("and"),
             VP(V("eat"),apple),
                VP(V("love"),apple.pro())))
\end{lstlisting}
\noindent
is realized as \realization{He\ eats\ an\ apple\ and\ loves\ it} which is expected.
But later in the program, the expression
\begin{lstlisting}
                S(apple,VP(V("be"),A("red")))
\end{lstlisting}
\noindent
will be realized as \realization{It is red.}, in which the pronominalization of \lstinline!apple! is still in effect. 

If this is not what was intended, then a new \lstinline!apple! object must be created before pronominalization. To achieve this, one can call \lstinline!clone()! which creates a new copy of the data structure. This is implemented by traversing the object and creating a string that corresponds to the \jsr{} expression for building this object. The resulting string is then evaluated in the current context to build a copy of the original expression. 

So our previous \lstinline!S! could have been coded as
\begin{lstlisting}
        S(Pro("him").c("nom"),
          CP(C("and"),
             VP(V("eat"),apple),
             VP(V("love"),apple.@\underline{clone()}!.pro())))
\end{lstlisting}
\noindent
after which
\begin{lstlisting}
        S(apple,VP(V("be"),A("red")))
\end{lstlisting}
\noindent
is realized as \realization{The\ apple\ is\ red.}

Seasoned \js{} programmers might prefer defining arrow functions instead of variables that will create a new data structure at each call, thus avoiding the need for cloning the original expression. For example, we might define
\begin{lstlisting}
        let apple = ()=>NP(D("a"),N("apple"))
\end{lstlisting}
\noindent
and call \lstinline!apple()! when needed. This also allows parametrizing the expression at each call.

\subsection{Formatting}
\label{formatting}
\jsr{} being aimed at web developers, it is important to deal with the generation of HTML tags using the method \lstinline!tag(name,attributes)! where \lstinline!attributes! is an optional object whose keys are attribute names and values are the corresponding attribute values. When this method is encountered, it saves the parameters and the values in the constituent. At \emph{stringification} time, these values are used to create the final string while taking into account HTML tags. For example

\begin{lstlisting}
S(Pro("him").c("nom"),
  VP(V("eat"),
     NP(D("a"),
        N("apple")
          .tag("a",
              {href:'https://en.wikipedia.org/wiki/Apple'}))
 )) 
\end{lstlisting}
\noindent
is realized as
\begin{center}\realization{
He eats an \textless{}a\ href="https://en.wikipedia.org/wiki/Apple"\textgreater{}apple\textless{}/a\textgreater{}.}
\end{center}
 
\noindent
in which the elision is performed between \realization{a} and \realization{apple} even though in the realized string the first letter of \lstinline!apple! does not appear immediately after \realization{a}, but it will be the case when it is displayed.

The punctuation before, after and around constituents is dealt similarly. The appropriate values of the strings to be inserted are saved within the constituent structure and used during the \emph{stringification} process.

Note that this implementation choice implies a \emph{small} limitation: HTML and other formatting can only be performed at constituent boundaries.

%

\subsection{Bilingual text generation} 
\label{sub:bilingual_text_generation}

As the \systeme{B} in its name implies, \jsr{} was designed for realizing bilingual texts (French and English) for which linguistic information differ: each language has its lexicon, its declension and conjugation rules, word ordering and even specialized formatting conventions. In \jsr{} each \lstinline!Terminal! carries an indication of the language to be used at realization time. This allows realizing a sentence in which the language of some words differ from the others. Although this feature is seldom used with in a sentence, it proved quite useful in some applications. In order to simplify the creation of a \lstinline!Terminal!, most calls are done in the context of the \emph{current language} which is set by calling \lstinline!loadEn()! for English or \lstinline!loadFr()! for French. These functions can be called as often as necessary even in the middle of a sentence.  For example:
\begin{lstlisting}
    loadFr();
    var dest=NP(D("le"),N("monde"));
    loadEn();
    S(Pro("I").pe(1),
      VP(V("say"),
         Q("hello"),
         PP(P("to"),dest.tag("b"))))
\end{lstlisting}
will be realized as \realization{I say hello to \textbf{le monde}.} 

\subsection{Other issues} 
\label{sub:other_issues}

\jsr{} having been designed for use in the context of data-to-text systems, it must deal with the proper generation of numbers. On top of the usual formatting, it also deals with the writing of numbers in letters while taking into account the number agreement for the noun: 

\lstinline!NP(NO(1).dOpt({nat:true}),N("plane"))! realized as \realization{one plane} while\\ \lstinline!NP(NO(3).dOpt({nat:true}),N("plane"))! is realized as \realization{three planes})

Number agreement for the noun phrase depends on the value of the number, note that rules for French and English slightly differ in this respect. It is also important to format dates in different forms: in letters or numbers, perhaps ignoring some components (e.g. show only date or time).  Relative dates (e.g. \realization{tomorrow, yesterday, last Monday,...}) can also be realized.  

\jsr{} also allows choosing randomly between a list of alternatives. This feature is useful for varying equivalent wordings for similar informations. It is used in a demonstration program for explaining subway routes between Metro stations of Montréal described in the next section. It could also be used to create random text generators in the spirit of Perchance.\footnote{\url{https://perchance.org/welcome}}.

\subsection{\pyr{}: Python implementation} 
\label{sub:python_implementation}

Given the fact that Python has become the \emph{lingua franca} of NLP over the last years, we decided to \emph{port} the \jsr{} \js{} code to Python called \pyr{}, while keeping the same constituency and dependency syntax notation\footnote{\pyr{} is available on \href{https://github.com/lapalme/jsRealB}{GitHub} with demonstration applications or as a Python package at \url{https://pypi.org/project/pyrealb/} }. As it is also available as a Python library that can be installed using \texttt{pip}, \pyr{} can then be embedded within Python applications by simply adding:

\begin{lstlisting}
        from pyrealb import *
\end{lstlisting}

Although the input notation for \jsr{} was designed to be \js{} expressions, given their similarity with \py{} expressions, the same notation is valid as a \py{} expression. In \js{}, the creation of constituents (e.g. \lstinline!S!, \lstinline!NP!, \lstinline!V!, ...) is done by means of function calls in upper case to mimic the traditional linguistic notation without having to add \texttt{new} in front of the constructor. In \py{}, constructors starting with a capital are simply called, so that \lstinline!Terminal!s such as \lstinline!D!, \lstinline!N! or \lstinline!V! and \lstinline!Phrase!s such as \lstinline!NP!, \lstinline!VP! and S are subclasses. Calls to functions and methods of objects in both languages use identical conventions.

Another difference is the fact, that \lstinline!false!, \lstinline!true! and \lstinline!null! are respectively \lstinline!False!, \lstinline!True! and \lstinline!None! in \py{}. In order to be able to use exactly the same notation in \py{} as in \js{},  the three  variables are defined as global in \py{} with the corresponding \py{} values. This means that all \jsr{} expressions can be used verbatim in \pyr{}.


\subsection{Implementation notes} 
\label{sub:implementation_notes}

An object-oriented approach was selected for organizing both \jsr{} and \pyr{}: 
\begin{description}
    \item[\texttt{Constituent}] : the main class defining common methods for all objects, such as access to properties, stringification, elision, formatting, etc.;
    \item[\texttt{Terminal}] : subclass of \lstinline!Constituent! that defines methods for creating words and for declension and conjugation;
    \item[\texttt{Phrase}] subclass of \lstinline!Constituent! that comprises methods for phrase creation and modifications;
    \item[\texttt{Dependent}] subclass of \lstinline!Constituent!\footnote{Some linguists might be offended by the fact that a \emph{dependency} is a kind of \emph{constituent}, but this organization simplified the implementation.} that defines methods for dependency creation and modifications;
\end{description}

\jsr{} realizes any legal expression even in the case of \emph{erroneous} specification by wrapping the terminal within double square brackets to indicate that the input should be checked. In such cases, it also writes on the \js{} console a warning explaining the error. This warning, using the language of the terminal in error, is realized by \jsr{} itself. This \emph{localization} is achieved by creating a table of \jsr{} functional expressions in French and English indexed by a warning indicator. Here is an excerpt of this table which contains 25 keys.
\begin{lstlisting}
Constituent.prototype.warnings = {
    "bad parameter":
        {en:(good,bad)=> 
            S(NP(D("the"),N("parameter")),
              VP(V("be").t("ps"),Q(good).a(","),Adv("not"),Q(bad))).typ({mod:"nece"}),
         fr:(good,bad)=> // le paramètre devrait être $good, pas $bad
            S(NP(D("le"),N("paramètre")),
              VP(V("être").t("c"),Q(good).a(","),Adv("non"),Q(bad))).typ({mod:"nece"})},
    ...
}
\end{lstlisting}
The call  \lstinline!warn("bad parameter","string","number")! when trying to create \lstinline!N(23)! will be realized as \realization{The parameter should be string, not number.} in English, but as \realization{Le paramètre devrait être string, non number.} in French.


\section{Applications} 
\label{sec:applications}
The \href{https://github.com/rali-udem/jsRealB\#demos}{\jsr{} GitHub demo repository} shows many examples of use of \jsr{} for some specific features: conjugation of any verb and declension of any word, showing all possible sentence modifications (e.g. negation, passivation, interrogation, etc.) (shown in Figure~\ref{fig:SentenceTypes}) from a single sentence structure, building a sentence structure using menus, a grammar game and the creation of a random sentence.

\jsr{} has also been used to create variations on existing texts such as \emph{Les exercices de style} of Raymond Queneau, also in its English version. It was also integrated in a web page for illustrating the flowchart for \emph{L'augmentation} of Georges Perec. The fact that it has been used to reproduce verbatim one version of the classical story of \emph{Little Riding Hood} in both French and English illustrates that the coverage of \jsr{} is quite extensive as it will be discussed in the next section.

\jsr{} has been used for Data to Text applications such as the \href{http://www.macs.hw.ac.uk/InteractionLab/E2E/}{E2E challenge} although \jsr{} was developed after the competition and thus too late for being part of the competition and be evaluated as such. There is also a demonstration of the description of a list of events given in a table; in fact this application was one of the first application developed with \jsr{} and served as an initial motivation.

\begin{figure}[h]
  \centering
    \includegraphics[width=.63\textwidth]{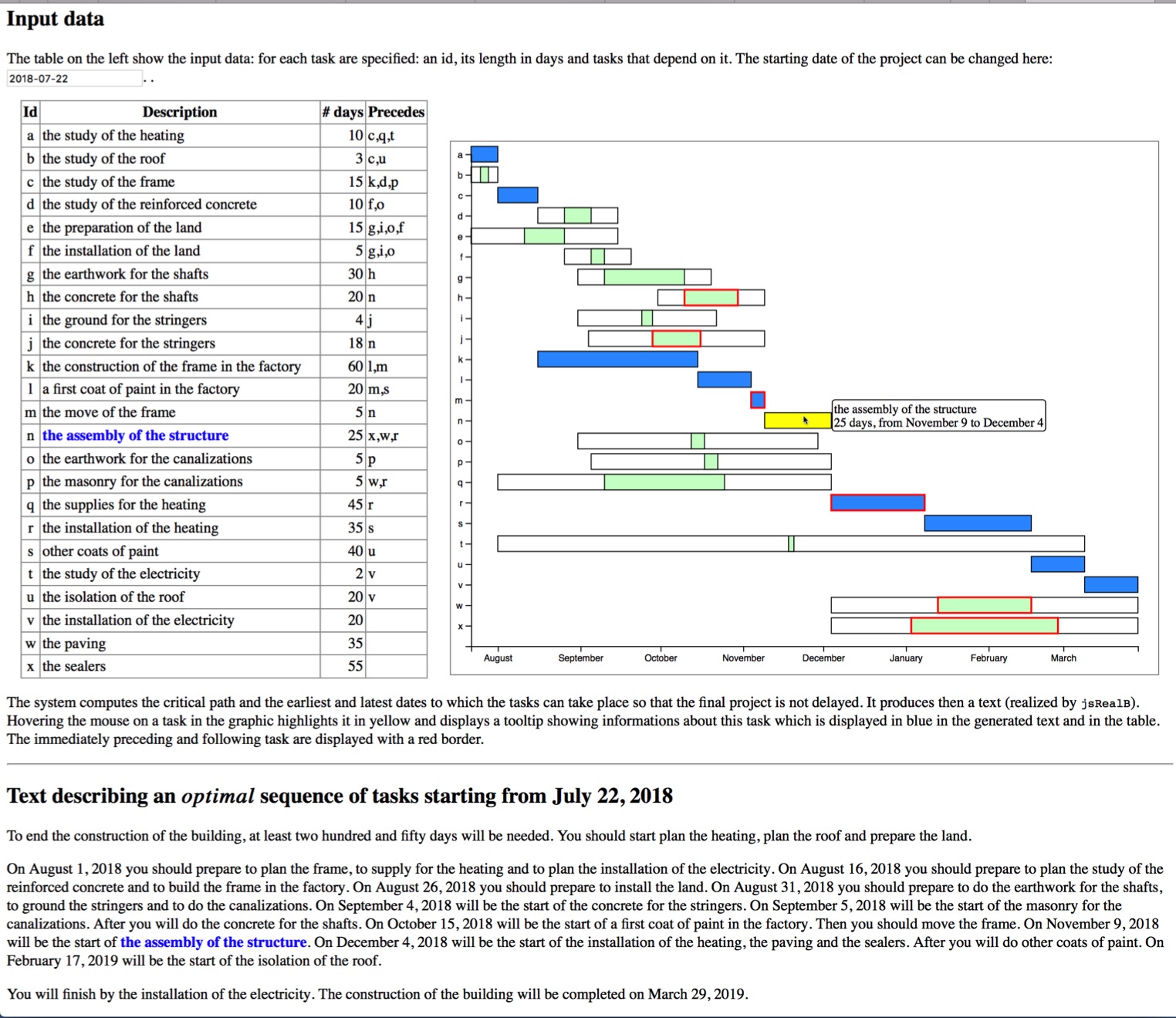}
    \hfill
    \includegraphics[width=.36\textwidth]{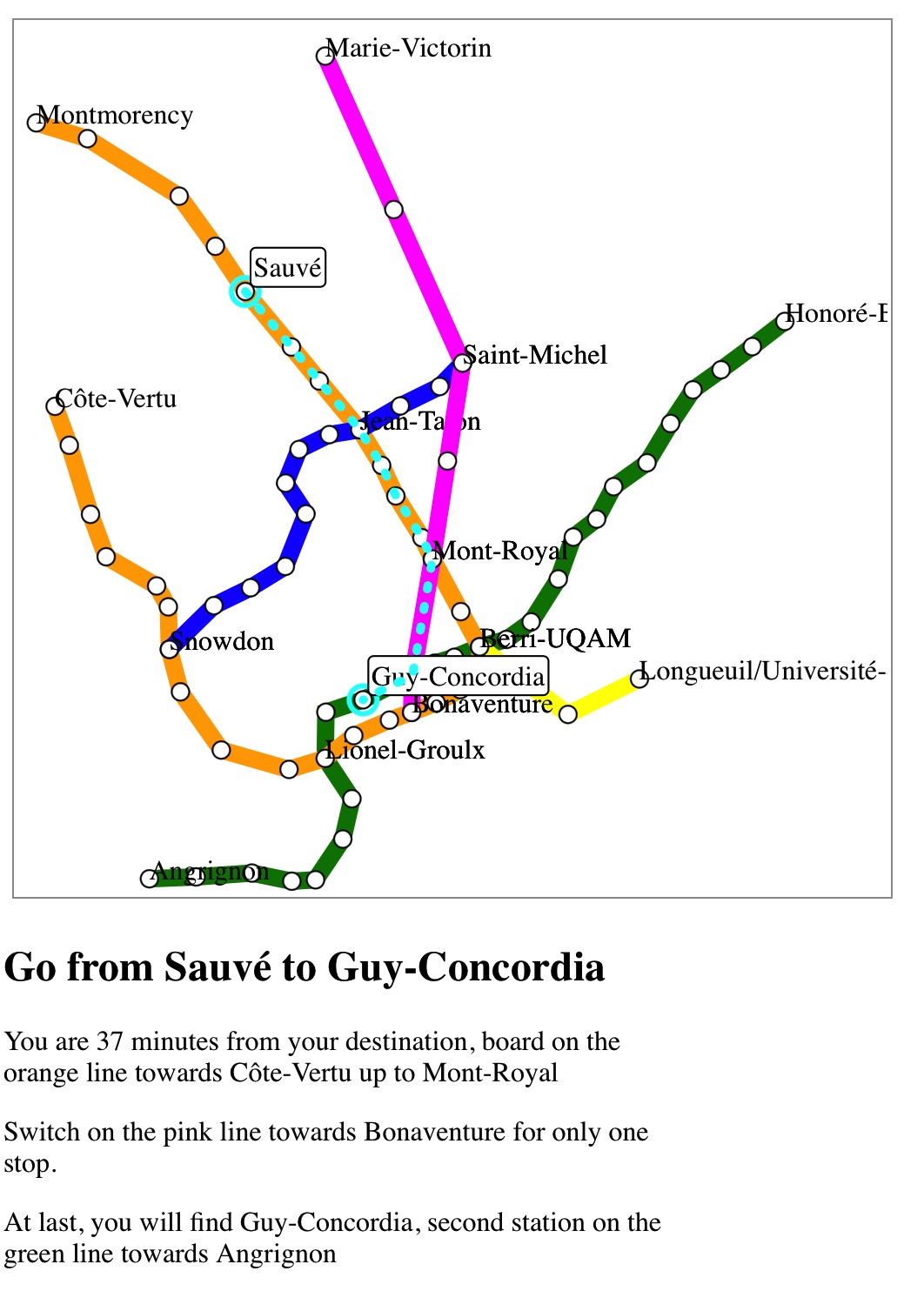}
  \caption{Screen shots of web based application with texts realized using \jsr{}. Oa n the le-t are described the steps for building a house after critical path computation, on the right is an interactive map of the Montréal metro which shows the itinerary between two stations.} 
  \label{fig:building-metro}
\end{figure}

T-o applications show text realization from data, but after non trivial computations in both French and English (see Figure~\ref{fig:building-metro})
\begin{itemize}
    \item The description of the building of a house given information about tasks, the duration and the precedence relations between them. The system first computes the critical path to find the start and end times of each task. It then creates a graphic for displaying the PERT diagram and an accompanying text to explain the steps to follow. It is possible to interactively change the start date and to explore the graphic with the mouse which also uses \jsr{} to generate the text of the tooltips.
    \item  The description of an itinerary in the Montréal Métro network. The system shows an interactive map of the Montréal Métro station. When a user clicks two stations, the systems realizes a text describing the itinerary to go from the first station to the second after having computed the shortest path between them.  
\end{itemize}

The GitHub repository also contains a prototype of a weather bulletin generation system\footnote{Available at \url{http://rali.iro.umontreal.ca/JSrealB/current/demos/Weather/Bulletin-generation.html}} that goes from numerical information encoded in JSON to well formed English and French bulletin. This application is particularly interesting to us because at one crucial time, Environment Canada provided some of the funding for the development of \jsr{}. This demonstration shows how the \emph{What to say} part of a natural language generation system written in \py{} can be linked with the \emph{How to say} in \js{} using the JSON notation of \jsr{}\footnote{This feature is not so useful now that \pyr{} is available but it remains an interesting exercise. In fact, it is this demonstration that prompted us to develop \pyr{}.}. 

\jsr{} has been embedded into an \href{https://observablehq.com/@lapalme/exprimenting-with-jsrealb}{Observable notebook} in which it is possible to change the existing expressions to see how their new realization. 

\jsr{} can also be used from a \emph{console} as a \systeme{node.js} modrule. An interactive IDE allows the testing of expressions, seaching ,for entries in the lexicon and also \emph{lemmatization}, i.e. given a form, show all possible \jsr{} expressions to realize it. For example, given \realization{springs}, it returns \texttt{N("spring").n("p")} and \texttt{V("spring")}. 

This \systeme{node.js} module can be used as a web server to realize sentences produced by a system written in another programming language. We have used it from a system written in Prolog and another in Python for developing symbolic realizers the context of recent NLG competitions: Surface Realization Shared Task 2019~\cite{Lapalme19SRST} and Web NLG challenge 2020~\cite{castro-ferreira20:_2020_bilin_bidir_webnl}.


\section{Coverage} 
\label{sec:coverage}

\subsection{Orthography} 
\label{sub:orthography}

\jsr{} can realize bilingual texts mixing both French and English because each word or phrase is associated with a specific language. It comes bundled with two relatively large lexica (52~560 entries for French and 33~933 for English) that cater for most uses: it has declension tables for all determiners, nouns, adjectives and pronouns in their tonic and clitic forms, it has conjugation tables for all English and French verbs even defective ones. 
On top of all simple tenses for French and English verbs for all persons and numbers, \jsr{} can also realize verbs with auxiliaries such as \emph{passé composé} or \emph{subjonctif plus-que-parfait} in French orthe  progressive or perfect verbs in English. It can also realize comparative and superlative forms of adjectives.

It is easy to add new words to the lexicon by specifying the appropriate table number. The IDE eases the finding the right table number, for example by providing entries with a given ending. 

\subsection{Syntax} 
\label{sub:syntax}
\jsr{} has all the necessary syntactic constructs to build any English and French sentence dealing automatically with most agreements between parts of a sentence (i.e., agreement between the subject and the verb or between different constituents with a noun phrase), a feature that is especially important in French. Most often, number and sometimes gender only need to be specified for nouns and tense for verbs, because other dependents agree automatically.


\section{Evaluation} 
\label{sec:evaluation}
As described in Section~\ref{sec:applications}, \jsr{} has been used quite extensively in many contexts: in NLG competitions, as the last step for an AMR realizer and in data-to-text applications. The system distribution comes with almost 8~000 unit tests: 6~000 for French conjugation and declension, 1~000 for English declension and conjugation; 100 for English pronouns, dates and numbers and 150 for French. There are also about 100 French and 50 English sentences that test specific \emph{difficult} features in sentences, such as agreement between multiple subjects, pronominalization and sentence transformations.  

Generating a sentence is \emph{instantaneous} needing only a few milliseconds on a commodity laptop; execution time has never been an issue with this realizer. For example, it takes less than 700  milliseconds to run the 8~000 unit tests.

We never encountered any serious limitation in generating text even when reproducing existing texts (e.g., \href{http://rali.iro.umontreal.ca/JSrealB/current/demos/PetitChaperonRouge/LittleRedRidingHood.html}{Little Riding Hood}), but we now describe a more formal evaluation by sampling sentences found in public corpora and reproducing them with \jsr{}, similarly to the methodology used by Braun \emph{et al.}~\cite{braun-etal-2019-simplenlg} for evaluating \snlg{}-DE. 

In order to find representative English and French sentences, we randomly sampled ten sentences of more than 5 tokens\footnote{We fell that there is no real NLG interest in generating very short sentences.} from the test corpora of each of the 6 English and 6 French Universal Dependencies 2.7 treebanks~\cite{UD-2.7} for which the morphological features are specified. This gave us 60 English sentences and 60 French sentences\footnote{As the following versions of UD (2.9 at the time of writing this revision) still contain the same sentences, perhaps with a few annotation corrections, we did not feel necessary to redo the sampling of sentences for these new versions. The goal was to reproduce the sentences verbatim and not according to their annotations.}. Using \lstinline!Q! (i.e., \emph{canned text}) only for specialized words which were rare, we created \jsr{} expressions to reproduce verbatim all sentences of our sample. This shows that the coverage of the French and English linguistic phenomena is almost complete. We only encountered a few limitations: \jsr{} cannot reproduce contracted forms such as \realization{I'll} for \realization{I will} or \realization{aint} for \realization{is not}, although there are provisions for contractions such as \realization{don't} or \realization{can't}; some word orders cannot be easily obtained such as inserting in adverb within a negation: \realization{You can also remove} will be realized as \realization{You also can remove}; possessive using \realization{'s} cannot be generated from the structure although it can be \emph{tricked} add the option \realization{.a("'s")} to the word referring to the possessor.

To \emph{regenerate} these 120 sentences, we could have written the \jsr{} expressions \emph{from scratch} using a program editor. It takes about two minutes to create an expression for a sentence of about 20 tokens using an \href{http://rali.iro.umontreal.ca/JSrealB/current/demos/Evaluation/index.html}{interactive editor embedded in a web page} which allows for testing the edited expression. 

But, for sentences already annotated with Universal Dependencies, we developed a tool for creating a \jsr{} expression using lemma, part of speech and features information. As the page highlights differences between the original UD text and the sentence realized by \jsr{}, it is a simple matter to edit the generated expression to reproduce the original text. To our dismay, we found many cases in which the original annotations were incorrect or incomplete: 25 sentences out of our sample of 60 English sentences (42\%) and 19 of 60 French sentences (31\%) had at least one discrepancy. This initial experiment, originally designed to measure the coverage of \jsr{} was then pursued with another goal, developing a tool to help validating UD annotations. This work is discussed in more detail in~\cite{UDregenerator2020Lapalme,lapalme-2021-validation}.  

\section{Conclusion}
\label{conclusion}

We have described the text realization process of \jsr{}, an interesting middle ground between a very abstract input specification and a detailed formatting language. It allows automating the \emph{finishing touches} for well-formed language strings displayed to a user.  We have shown that the seemingly simple task of producing well-formed natural language text from a relatively abstract formalism involves a lot of intricate language dependent details that must be dealt with, for realizing fluent and syntactically correct sentences.

\jsr{} covers the most important basic features of both French and English and comes with comprehensive lexica covering most current uses  for which it is straightforward to add new words.
The implementation was validated by testing for grammatical functionality, e.g., verb conjugation, and language coverage on sentences taken from UD dependencies. This approach is also available as a Python package.

\section*{Acknowledgments} 
\label{sec:acknowledgments}
\jsr{} was originally developed for French by Nicolas Daoust during his master's thesis. The bilingual version was then created by Paul Molins and improved by Francis Gauthier during their internship at RALI. Over the years, it has benefited from suggestions by members of the RALI, especially Fabrizio Gotti. We also thank François Lareau from OLST in the Linguistics Department of Université de Montréal for many fruitful discussions and suggestions.
\clearpage
\bibliographystyle{plain}
\bibliography{jsRealB-Architecture.bib} 

\end{document}